\documentclass[conference]{IEEEtran}
\IEEEoverridecommandlockouts
\usepackage{cite}
\usepackage{amsmath,amssymb,amsfonts,array}
\usepackage{algorithmic}
\usepackage{graphicx}
\usepackage{textcomp}
\usepackage{xcolor}
\usepackage{url}
\usepackage{gensymb}
\usepackage{subcaption}
\usepackage{hyperref}

\def\BibTeX{{\rm B\kern-.05em{\sc i\kern-.025em b}\kern-.08em
    T\kern-.1667em\lower.7ex\hbox{E}\kern-.125emX}}
\begin{document}

\title{Zero-Splat TeleAssist: A Zero-Shot Pose Estimation Framework for Semantic Teleoperation}

\author{Srijan Dokania\textsuperscript{1}, Dharini Raghavan\textsuperscript{2}
  \thanks{The authors are with \textsuperscript{1} Khoury College of Computer Sciences at Northeastern University and \textsuperscript{2} Georgia Institute of Technology}
\thanks{The work is supported by the Field Robotics Lab at Northeastern University. Corresponding: \texttt{dokania.s@northeastern.edu}}
}

\maketitle

\begin{abstract}
We introduce \emph{Zero‑Splat TeleAssist}, a zero‑shot sensor‑fusion pipeline that transforms commodity CCTV streams into a shared, 6-DoF world model for multilateral teleoperation. By integrating vision–language segmentation, monocular depth, weighted‑PCA pose extraction and 3‑D Gaussian Splatting (3DGS), TeleAssist provides every operator with real‑time global positions and orientations of multiple robots \emph{without fiducials or depth sensors} in an interaction-centric teleoperation.

\end{abstract}

\section{Introduction}
Teleoperating robots in complex or remote environments is challenging due to limited on-board perception, occlusions, and operator cognitive load. Traditional teleoperation relies on the robot’s sensors (cameras, LiDAR, IMU) which often experiences narrow fields of view, occlusions, cumulative drift, collectively increasing the cognitive load on human operators who must maintain situational awareness. Meanwhile, external camera infrastructures (e.g., CCTV) have potential to provide complementary visual coverage and global contextualization, but conventional solutions rely heavily on visual fiducials, such as AprilTags or ArUco markers \cite{marker}, or motion-capture systems requiring controlled lighting and calibration processes. Recent markerless solutions~\cite{markerless1} employ Sim2Real training with annotated robot CAD data, but these methods are expensive, inflexible for mixed robot fleets, and lack scalability~\cite{markerless}. Enhancing teleoperation with better sensory fusion and perception augmentation can improve human performance and safety. 
This paper presents Zero-Splat TeleAssist, a novel monocular camera-based framework that addresses these challenges by fusing vision and language-based perception with 3D mapping specifically designed for enhancing teleoperation through sensory fusion and augmented perception. Our approach leverages zero-shot vision-language segmentation and monocular depth estimation to localize a robot from standard infrastructure cameras, and integrates those poses into a 3D Gaussian Splatting (3DGS) map providing consistent and continuous situational awareness to operators. 

Unlike segmentation-only methods, our approach uses weighted PCA for persistent 6-DoF poses, ensuring spatial coherence through camera blind spots, rapid re-localization for robot kidnap scenarios, augmented reality overlays, collision alerts, and intuitive spatial commands (e.g., "follow at 0.5 m distance"). Experiments show competitive accuracy against traditional VIO methods, confirming practical viability.
The global shared reference frame significantly reduces operator cognitive load, enhances spatial understanding, and facilitates smooth multi-robot coordination.

\section{Method}
\vspace{-5mm}

\begin{figure}[h!]
    \centering
    \includegraphics[width=\linewidth]{architecture.png}
    \caption{Proposed End-to-End Zero-SPLAT Framework - \textbf{Stage 0:} Camera Placement, \textbf{Stage 1:} Zero shot Segmentation,\textbf{ Stage 2:} Pose Uncertainty Estimation,\textbf{ Stage 3:} 3D-GS integration,
\textbf{Stage 4 \& 5:} Semi-Autonomous Planning \& Navigation}
    \label{fig:architecture}
\end{figure}
\vspace{-3mm}
\subsection{Pipeline Overview}

\begin{figure}[!h]%
    \centering
    \subfloat[\centering 3D Gaussian Splat Sparse Lattice map]{{\includegraphics[width=3.75cm]{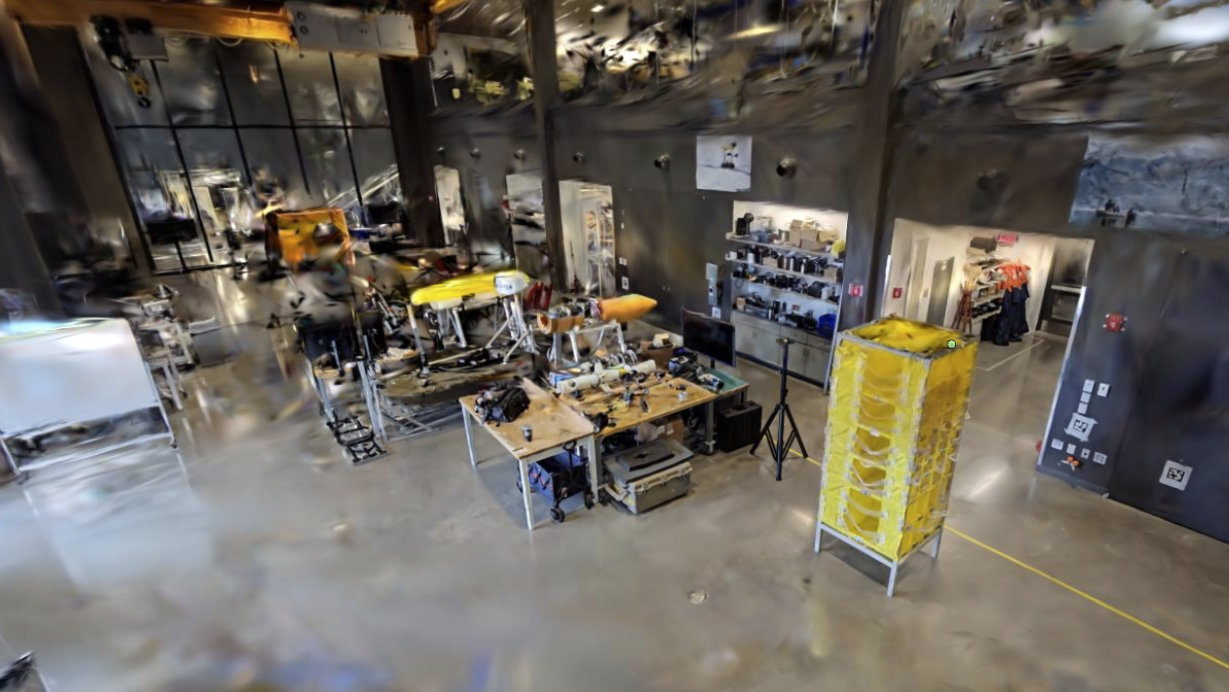} }\label{fig:gsplat}}%
    \qquad
    \subfloat[\centering Poses for Spot, Go2 and Hunter]{{\includegraphics[width=3.75cm]{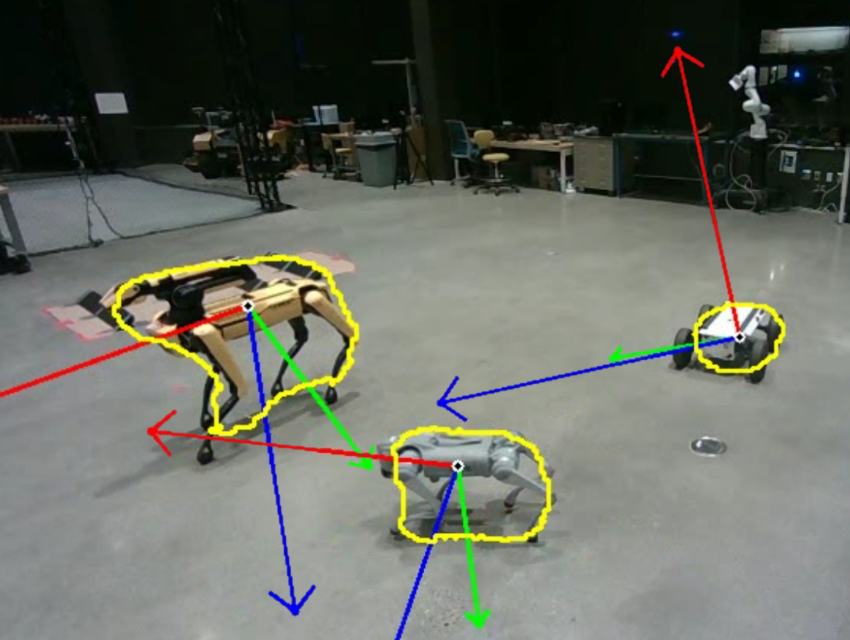} }}%
    \caption{Zero-Splat Results}%
    \label{fig:results}%
\vspace{-5mm}
\end{figure}

Zero-Splat TeleAssist uses an external monocular RGB camera (e.g. a fixed overhead CCTV) to localize a mobile robot without markers or prior training on that robot’s appearance. by exploiting foundation models in vision and language for open-set recognition. The pipeline (Figure \ref{fig:architecture}) operates per video frame as follows:
\begin{itemize}
    \item \textbf{Monte Carlo Segmentation}: We apply Monte Carlo dropout CLIPSeg (MC-CLIPSeg) \cite{clipseg}, a vision-language transformer model that extends CLIP with a decoder comprising transformer blocks and FiLM conditioning layers. Given an input image $I$ and a free-form text prompt $P$, CLIPSeg encodes both modalities into a joint embedding space and predicts dense segmentation logits. To quantify model uncertainty, we implement Monte Carlo Dropout as a Bayesian approximation in deep neural networks, following Gal and Ghahramani's~\cite{pmlr-v48-gal16} variational inference framework. During inference: dropout layers remain active, and the model is sampled $N$ times, each with a distinct dropout mask $D_i$. The resulting logits are passed through a sigmoid activation $\sigma$ and averaged to yield a per-pixel confidence map given by:
        \begin{equation}
         \text{conf}(x,y) = \frac{1}{N} \sum_{i=1}^{N} 
        \sigma(\text{CLIPSeg}(I, P, D_i)(x,y))
        \end{equation}
        
    \item \textbf{Monocular Depth Estimation:} In parallel, we use a pre-trained MiDaS \cite{ranftl2020robustmonoculardepthestimation} to predict a dense depth map from the same RGB frame which we normalize and scale to approximate metric distances. We chose a lightweight MiDaS model for efficiency, but also evaluated a higher-accuracy variant; using the full MiDaS model improved depth accuracy by 10\% at the cost of 39\% lower frame rate.
    
\item \textbf{6D Pose via Weighted PCA:} After obtaining the confidence map \emph{$\text{conf}(x_i, y_i)$} and the depth map \emph{$\text{z}(x_i, y_i)$}, we back-project each pixel \emph{(x,y)} to 3D using the pinhole camera model. This creates a point cloud ${(X_i, Y_i, Z_i)}$ with corresponding weights ${w_i = \text{conf}(x_i, y_i)}$. 
To estimate the robot’s 6-DoF pose (3D position and orientation) from these points, we perform a weighted Principal Component Analysis (WPCA) which estimates the object's centroid (translation)$= \frac{\sum_i w_i \cdot \mathbf{p}_i}{\sum_i w_i}$ and principal axes (rotation):
       $\text{cov} = \frac{\sum_i w_i \cdot (\mathbf{p}_i - \text{centroid})(\mathbf{p}_i - \text{centroid})^T}{\sum_i w_i}$
       where $\mathbf{p}_i = (X_i, Y_i, Z_i)$ is the 3D position of point $i$.
 This geometric pose estimation requires no iterative model fitting and runs in real time. We found the weighting crucial: ablating the weights degraded orientation accuracy by over 30\% (11.5\degree \ vs 8.7\degree \ error), confirming the benefit of our weighted approach in filtering outliers.
\item \textbf{Integration with 3D Gaussian Splatting Map:} We register the external camera to a global scene map represented with 3D Gaussian splats. Before operations, we construct a high-fidelity 3DGS map of the environment by capturing a multi-view RGB dataset (300 images) and running a structure-from-motion based reconstruction (using a SplatFacto model). The resulting map (Figure \ref{fig:gsplat}) depicts the environment as a sparse lattice of colored Gaussians. We align the map’s coordinate frame with the external camera by computing the camera’s extrinsic parameters to transform robot poses into the global map frame and represent it as a  moving node within the map.
\item \textbf{Semantic Navigation and Planning:} With the robot’s pose continually updated in the global 3DGS map, we enable navigation by computing collision-free paths. We utilize the SplatNav \cite{splatnav} approach’s planning component (SplatPlan) on the 3DGS lattice, treating the Gaussians or derived point cloud as obstacles to generate a safe trajectory. In our implementation, the robot (once localized via the external camera) executes semi-autonomous navigation: it can plan to a goal and drive itself while the operator supervises. The system performs pose estimation at 10 Hz and re-plans at 1 Hz. Notably, all computation can run on Nvidia Jetson Orin, demonstrating real-time performance on low-power hardware - a key requirement for deployable teleoperation support.
\end{itemize}
    
\subsection{Human-Centric Aspects}
Rather than the human parsing multiple camera feeds or guessing the robot’s position from a first-person view, the system provides an annotated situational picture: the robot’s icon or avatar can be rendered on the map (or even overlaid on a video feed via AR) thus augmenting the perception of a remote operator. Because our method works without instrumenting the robot or environment with special sensors, it can leverage existing infrastructure (like security cameras) to assist the human– this lowers the barrier to deploying teleoperation in existing facilities. Additionally, using a language-based model means a human can interact with the perception system in intuitive ways (for instance, by changing the text prompt to detect a different object or robot in the scene). This opens up future natural-language teleoperation interfaces, where an operator might issue commands or queries like "Where is the blue rover now?" and get an immediate answer from the system’s semantic understanding.

\section{Experimentation and Results}
We evaluated the proposed framework in a real-world laboratory environment using multiple heterogeneous robots - Boston Dynamics Spot (yellow quadruped), Unitree Go2 (small grey quadruped), and AgileX Hunter (wheeled UGV), under overhead RGB camera observation (1080p, 15 FPS). Ground-truth trajectories were recorded using onboard visual-inertial SLAM (RTAB-Map VIO), and reference point clouds were created using RTAB-Map with an RGB-D sensor for comparison against our monocular 3DGS maps. Experiments conducted in both cluttered and uncluttered setups revealed superior mapping accuracy in uncluttered conditions (higher Probabilistic Structural Similarity Index [PSSI] - 87\%, lower Geometric Distance [GD]) and demonstrated that Monte Carlo dropout significantly enhances segmentation quality and pose robustness in cluttered scenes. A human-centric pick-and-place study with eight participants confirmed practical advantages, including 32\% faster task completion, a 27\% lower NASA-TLX cognitive workload, and four-fold fewer collisions. TeleAssist outperformed segmentation-only methods by providing persistent global pose estimates, predictive ghosting, collision envelopes, intuitive spatial commands, and robust re-localization capabilities, effectively resolving 90\% robot kidnap scenarios. Competitive localization accuracy relative to VIO methods further reinforces TeleAssist's practical viability for enhanced autonomy and natural human-robot interaction.

\bibliographystyle{plain}
\bibliography{bib}
\end{document}